\documentclass[runningheads]{llncs}
\usepackage{graphicx}

\usepackage[table]{xcolor}
\usepackage{tikz}
\usepackage{comment}
\usepackage{multirow}
\usepackage{amsmath,amssymb} 
\usepackage{color}
\definecolor{Gray}{gray}{0.9}
\usepackage{subcaption}
\usepackage{array, caption, floatrow, tabularx, makecell, booktabs}%
\usepackage[accsupp]{axessibility}  

\def\etal{\emph{et al.}}

\newcommand{\ie}{\textit{i}.\textit{e}., }
\newcommand{\tp}{\textsuperscript}

\begin{document}
\pagestyle{headings}
\mainmatter
\def\ECCVSubNumber{6380}  

\title{ML-BPM: Multi-teacher Learning with Bidirectional Photometric Mixing for Open Compound Domain Adaptation in Semantic Segmentation} 

\titlerunning{ML-BPM}
%
\author{Fei Pan\inst{1} \and
Sungsu Hur\inst{1} \and
Seokju Lee\inst{2} \and
Junsik Kim\inst{3} \and
In So Kweon\inst{1}
}
\authorrunning{F. Pan et al.}
%
\institute{KAIST, South Korea. \email{\{feipan, sshuh1215, iskweon77\}@kaist.ac.kr}\and
KENTECH, South Korea. \email{slee@kentech.ac.kr} \and
Harvard University, USA. \email{mibastro@gmail.com}}
\maketitle

\begin{abstract}
Open compound domain adaptation (OCDA) considers the target domain as the compound of multiple unknown homogeneous subdomains. The goal of OCDA is to minimize the domain gap between the labeled source domain and the unlabeled compound target domain, which benefits the model generalization to the unseen domains. Current OCDA for semantic segmentation methods adopt manual domain separation and employ a single model to simultaneously adapt to all the target subdomains. However, adapting to a target subdomain might hinder the model from adapting to other dissimilar target subdomains, which leads to limited performance. In this work, we introduce a multi-teacher framework with bidirectional photometric mixing to separately adapt to every target subdomain. First, we present an automatic domain separation to find the optimal number of subdomains. On this basis, we propose a multi-teacher framework in which each teacher model uses bidirectional photometric mixing to adapt to one target subdomain. Furthermore, we conduct an adaptive distillation to learn a student model and apply consistency regularization to improve the student generalization. Experimental results on benchmark datasets show the efficacy of the proposed approach for both the compound domain and the open domains against existing state-of-the-art approaches.

\end{abstract}

\section{Introduction}
Semantic segmentation is a fundamental task in finding applications to many problems, including robotics~\cite{zhang2018shufflenet}, autonomous driving~\cite{zhao2017pyramid}, and medical diagnosis~\cite{ouyang2020self}.
Recently, deep learning-based semantic segmentation approaches~\cite{huang2019ccnet,zheng2021rethinking,zhao2017pyramid} have achieved remarkable progress. However, their effectiveness and generalization ability require a large amount of pixel-wised annotated data which are expensive to collect. To reduce the cost of data collection and annotation, numerous synthetic datasets have been proposed~\cite{richter2016playing,RosCVPR16}. However, the models trained on synthetic data tend to poorly generalize to real images. To cope with this issue, unsupervised domain adaptation (UDA) methods~\cite{Tsai_adaptseg_2018,vu2018advent,zou2018unsupervised,pan2020unsupervised,tranheden2021dacs,zhang2021prototypical} have proposed to align the domain gap between the source and the target domain. Despite the efficacy of UDA techniques, most of these works rely on the strong assumption that the target data is composed of a single homogeneous domain. This assumption is often violated in real-world scenarios. As an illustration in autonomous driving, the target data will likely be composed of various subdomains such as night, snow, rain, etc. Therefore, directly applying the current UDA approaches to these target data might deliver limited performance. This paper focuses on the challenging problem of open compound domain adaptation (OCDA) in semantic segmentation where the target domain is unlabeled and contains multiple homogeneous subdomains.  The goal of OCDA is to adapt a model to a compound target domain and to further enhance the model generalization to the unseen domains.\\
\indent To perform OCDA, Liu~\etal~\cite{liu2020open} propose an easy-to-hard curriculum learning strategy, where samples closer to the source domain will be chosen first for adaptation. However, it does not fully take advantage of the subdomain boundaries information in the compound target domain. To explicitly consider this information, current OCDA works~\cite{gong2021cluster,park2021discover} propose to separate the target compound domain into multiple subdomains based on image style information. Existing works use a manual domain separation method; they also employ a single model to simultaneously adapt to all the target subdomain. However, adapting to a target subdomain might hinder the model from adapting to other dissimilar target subdomains, which leads to limited performance. We propose a multi-teacher framework with bidirectional photometric mixing for open compound domain adaptation in semantic segmentation to tackle this issue. First, we propose automatic domain separation to find the optimal number of subdomains and split the target compound domain. Then, we present a multi-teacher framework in which each teacher model uses bidirectional photometric mixing to adapt to one target subdomain. On this basis, we conduct adaptive distillation to learn a student model and apply a fast and short online updating using consistency regularization to improve the student's generalization to the open domains. We evaluate our approach on the benchmark datasets. The proposed approach outperforms all the existing state-of-the-art OCDA techniques and the latest UDA techniques for domain adaptation and domain generalization task.\\
\noindent \textbf{The Contribution of This Work.} (1) we propose automatic domain separation to find the optimal number of target subdomains; (2) we present a multi-teacher framework with bidirectional photometric mixing to reduce the domain gaps between the source domain and every target subdomain separately; (3) we further conduct an adaptive distillation to learn a student model and apply consistency regularization to improve the student generalization to the open domains.

\section{Related Work}
\noindent \textbf{Unsupervised Domain Adaptation}. Unsupervised domain adaptation (UDA) techniques are used to reduce the expensive cost of pixel-wise labeling tasks like semantic segmentation. In UDA, adversarial learning is used actively to align input-level style using image translation, feature distribution, or structured output~\cite{tsai2018learning,hoffman2018cycada,vu2018advent,pan2020unsupervised,wang2020differential}. Alternatively, self-training approaches~\cite{araslanov2021self,zhang2021prototypical,tranheden2021dacs,zou2018unsupervised} have also recently demonstrated compelling performance in this context. While these works have shown significant improvement, adopting those works directly for practical usage shows limitations due to its restricted setting dealing with only single source and single target. Despite the improvement provided by UDA techniques, their applicability to real scenarios remains restricted by the implicit assumption that the target data contains images from a single distribution.

\noindent \textbf{Domain Generalization}. The purpose of domain generalization (DG) is to train a model -- solely using source domain data -- such that it can perform reliable predictions on unseen domain. While DG is an essential problem, a few works have attempted to address this problem in the task of semantic segmentation. DG for semantic segmentation shows two main streams: augmentation-based and network-based approaches. The augmentation-based approaches~\cite{yue2019domain,huang2021fsdr} propose to significantly augment the training data via an additional style dataset to learn domain-invariant representation. The network-based approaches~\cite{pan2018two,choi2021robustnet} attempt to modify the structure of the network to minimize domain-specific information (such as colors or styles) such that the resulting model mainly focuses on the content-specific information. Even though DG for semantic segmentation has achieve obvious progress, their performance is inevitably lower than several UDA methods due to the absence of the target images, which is capable of providing abundant domain-specific information.

\noindent \textbf{Open Compound Domain Adaptation}. Liu~\etal~\cite{liu2020open} firstly suggests Open Compound Domain Adaptation (OCDA) that handles unlabeled compound heterogeneous target domain and unseen open domain. While Liu~\etal~\cite{liu2020open} propose a curriculum learning strategy, it fails to consider the specific information of each target subdomain. Current OCDA works~\cite{gong2021cluster,park2021discover}
propose to separate the compound target domain into multiple subdomains to handle the intra-domain gaps. Gong~\etal~\cite{gong2021cluster} adopt domain-specific batch normalization for adaptation. Park~\etal~\cite{park2021discover} utilize GAN-based image translation and adversarial training to exploit domain invariant features from multiple subdomains.

\section{Generating Optimal Subdomains}
\begin{figure}[t!]
    \centering
    \includegraphics[width=\textwidth]{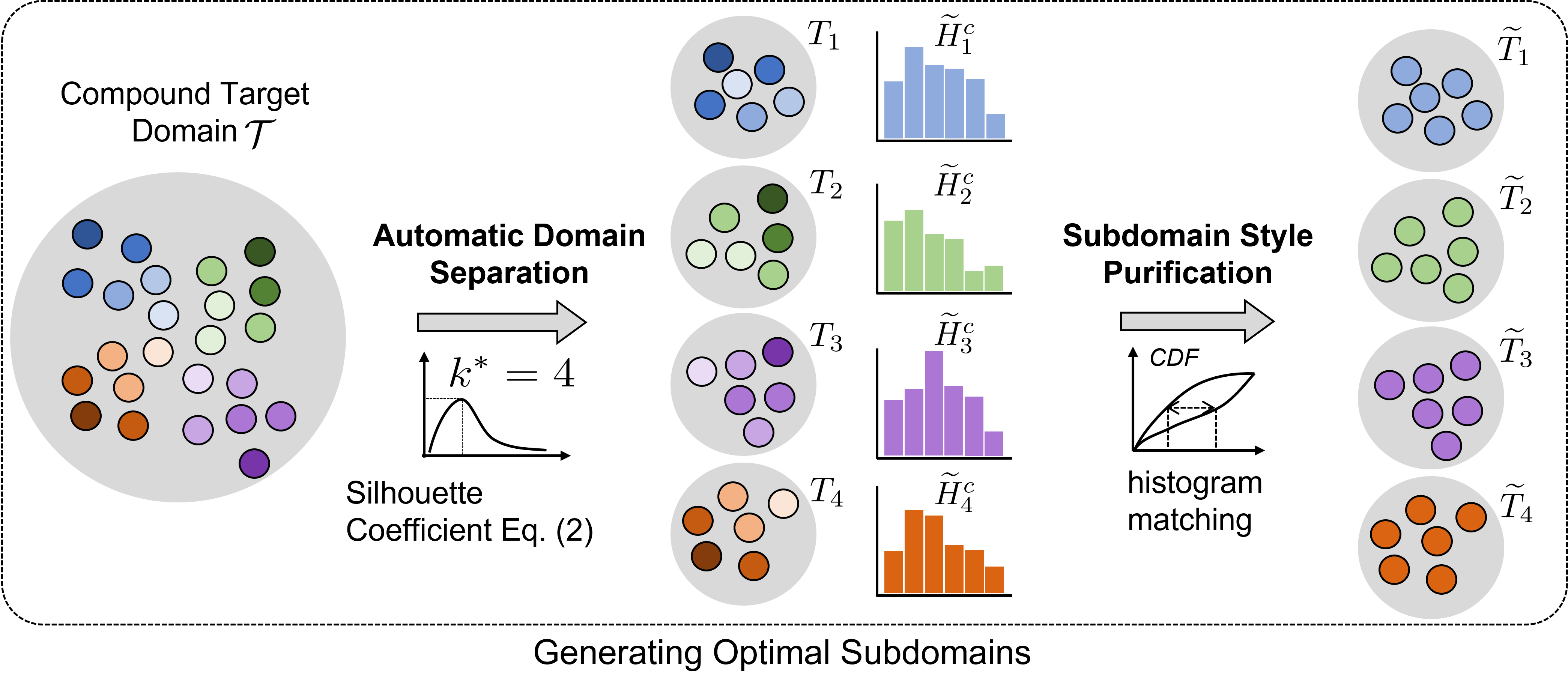}
    \caption{The part of generating optimal subdomains consists of automatic domain separation (ADS) and subdomain style purification (SSP). In ADS, we adopt Silhouette Coefficient~\cite{rousseeuw1987silhouettes} to find the optimal number of subdomains $k^{*}$. In SSP, we calculate mean of histogram $\widetilde{H}^{c}_m$ for the $m\tp{th}$ target subdomain $T_m$ according to Equation~\ref{eq:meanH}, and the purified subdomain is denoted as $\widetilde{T}_m$.}
    \label{fig:my_label}
\end{figure}
\subsection{Automatic Domain Separation}\label{ADS}
Our work assumes that the domain-specific property of images comes from their styles. Existing works adopt a predefined parameter to decide the number of subdomains, which might lead to a nonoptimal domain adaptation performance; furthermore, they rely on a pre-trained CNN-based encoder to extract the style information for the subdomain discovery. However, we propose an automatic domain separation (ADS) to effectively separate the target domain using the distribution of pixel values of the target images. The proposed ADS is capable of predicting the optimal number of subdomains without relying on any predefined parameters and extracting the image style information without relying on any pre-trained CNN models. We denote the source domain as $\mathcal{S}$, and the unlabeled compound target domain as $\mathcal{T}$. We also assume compound target domain contains $k$ latent subdomains: $\{T_1, \dots, T_k\}$, which lack of clear prior knowledge to distinguish themselves. The goal of ADS is to find the optimal number of subdomains $k^{*}$ and separate $\mathcal{T}$ into several subdomains accordingly.

Current work~\cite{ma2021coarse} suggests a simple yet effective style translation method by matching the distribution of pixel values on LAB color space. Thus, we adopt LAB space into ADS to extract the style information of the target image. Given a target RGB image $x_t \in \mathcal{T}$ as input, we convert it into LAB color space $rgb2lab(x_t)$. The three channels in LAB color space are represented as $l, a$, and $b$. Then, we compute the histograms of the pixel values for all three channels in LAB color space: $H^l(x_t), H^a(x_t)$, and $H^b(x_t)$. The histograms are concatenated and represented as the style information of $x_t$. Let $s(x_t)={H^l(x_t)}^\frown{H^a(x_t)}^\frown{H^b(x_t)}$ denote the concatenated histograms of $x_t$, and we take $s(x_t)$ as input to ADS for domain separation. However, most existing clustering algorithms require a hyperparameter to determine the number of clusters. Directly applying a naive clustering might lead to a nonoptimal adaptation performance. Thus, we propose to find the optimal number $k^*$ of the subdomains using Silhouette Coefficient (SC)~\cite{rousseeuw1987silhouettes}. Suppose the target domain $\mathcal{T}$ is separated into $k$ subdomains, $\{T_1, \dots, T_k\}$. For each target image $x_t$, we denote $\gamma(x_t)$ as the average distance between $x_t$ and all other target images in the target subdomain to which $x_t$ belongs. Additionally, we use $\delta(x_t)$ to represent the minimum average distance from $x_t$ to all other target subdomains to which $x_t$ does not belong. Let us assume $x_t$ belongs to the $m\tp{th}$ target subdomain $T_m$, then $\gamma(x_t)$ and $\delta(x_t)$ are written as
\begin{equation}
    \begin{aligned}
        \gamma(x_t) &= \frac{\sum_{{x_t}' \in T_m, {x_t}' \not = x_t}L( s({x_t}'), s(x_t) )}{|T_m |-1},\\
        \delta(x_t) &= \min_{T_n:1\leq n \leq k, n\not=m} \left \{ \frac{\sum_{{x_t}' \in T_n} L( s({x_t}'), s(x_t) )}{|T_n|} \right \},
    \end{aligned}
\end{equation}
where $L( s({x_t}'), s(x_t) )$ represents the euclidean distance of $s({x_t}')$ and $s(x_t)$, and $|T_m|$ is the number of the target images in $T_m$. 
The SC score for $k$ number of the target subdomains is given by
\begin{equation}
    SC(k) = \sum_{x_t\in \mathcal{T}} { \frac{\delta(x_t) - \gamma(x_t)}{\max (\gamma(x_t), \delta(x_t))}}.
\end{equation}
Hence, the goal of the proposed ADS is to find $k^*$ for
\begin{equation}
    k^* = \arg \max_{k} SC(k).
\end{equation}

\subsection{Subdomain Style Purification}\label{SSP}
With the help fo automatic domain separation, the number of abnormal samples with different styles is small inside each target subdomain. Though these abnormal samples might be useful for the model's generalization, they could also lead to a negative transfer, which further hinders the model from learning domain invariant features in a specific subdomain. To cope with it, we propose to purify the style distribution of the target images inside each subdomain. We design a subdomain style purification (SSP) module to effectively make similar styles for the images within the same subdomain. Given the $m\tp{th}$ target subdomain $T_m$, we adopt the histograms of LAB color space $\{(H^{l}(x_t),H^{a}(x_t),H^{b}(x_t)); \forall x_t \in T_m \}$ (mentioned in~\ref{ADS}), and then we compute the mean of the histograms for all the three channels, represented by $\widetilde{H}_m^l, \widetilde{H}_m^a$, and $\widetilde{H}_m^b$, and this process is achieved by
\begin{equation}\label{eq:meanH}
    \widetilde{H}_m^c = \frac{\sum_{x_t \in T_m} H^c(x_t)}{|T_m|}; \forall c \in \{l,a,b\},
\end{equation}
where $|T_m|$ represents the number of the target images in $T_m$. We take $\{\widetilde{H}_m^l, \widetilde{H}_m^a, \widetilde{H}_m^b \}$ as the standard style for $T_m$. For each target RGB image $x_t \in T_m$, we change the style of $x_t$ to generate the RGB new image $\tilde{x}_t$ by the histogram matching~\cite{rafael2010digital} on $\widetilde{H}_m^l, \widetilde{H}_m^a$, and $\widetilde{H}_m^b$ on the LAB color space. The process of SSP is done for all the subdomains $\{T_1, \dots, T_{k^*}\}$. We denote the purified subdomains after SSP as $\{ \widetilde{T}_1, \dots, \widetilde{T}_{k^*} \}$. 
\section{Multi-teacher Framework}
\begin{figure}[t!]
    \centering
    \includegraphics[width=\textwidth]{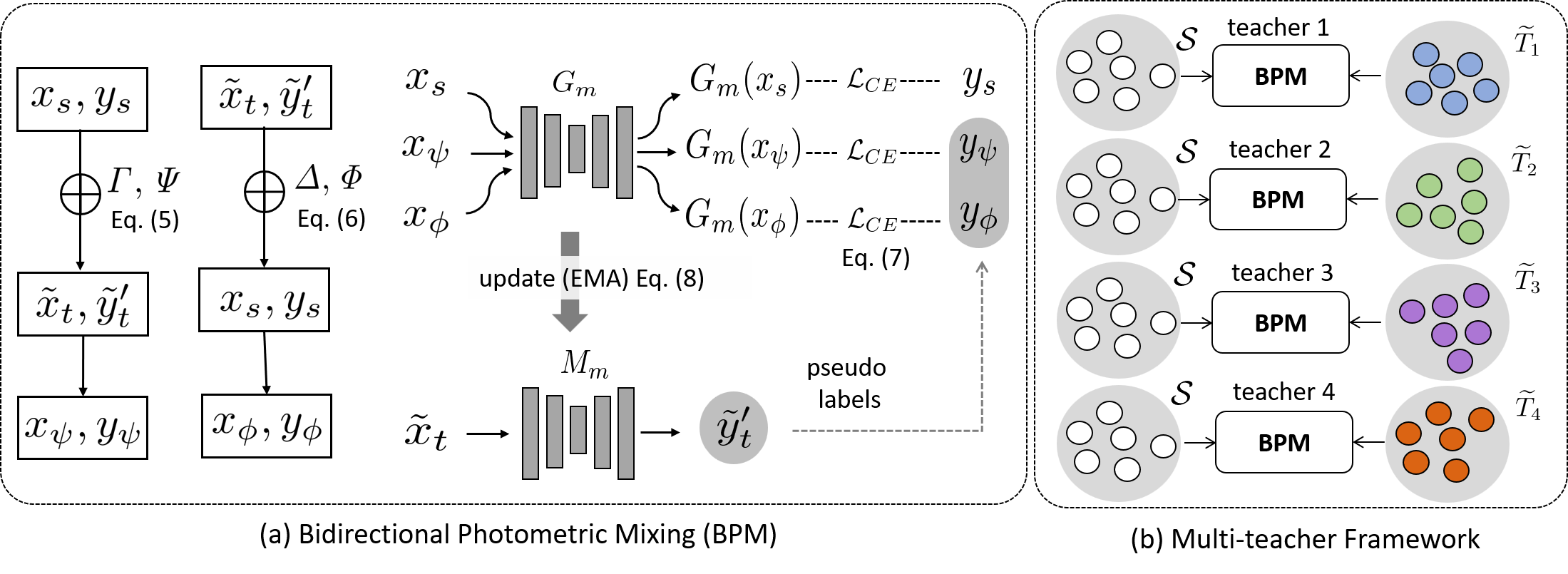}
    \caption{(a) The architecture of the proposed bidirectional photometric mixing. (b) The diagram of the multi-teacher learning framework.}
    \label{figure:BPM}
\end{figure}
\subsection{Bidirectional Photometric Mixing} 
Through automatic domain separation and subdomain style purification (mentioned in~\ref{ADS} and~\ref{SSP}), the compound domain $\mathcal{T}$ is automatically separated into multiple subdomains $\{ \widetilde{T}_1, \dots, \widetilde{T}_{k^*} \}$, where $k^*$ represents the optimal number of the subdomains. Our next plan is to minimize the domain gap between the source domain and each target subdomain. A recent UDA work DACS~\cite{tranheden2021dacs} presents a mixing-based UDA technique for semantic segmentation. Inspired by DACS, we propose bidirectional photometric mixing (BPM) to minimize the domain gap between the source domain and each target subdomain separately. Compared with DACS, the proposed BPM adopts a photometric transform to decrease the style inconsistency of the mixed images to reduce the pixel-level domain gap. On this basis, BPM applies a bidirectional mixing scheme to provide a more robust regularization for training. The architecture of BPM is shown in Figure~\ref{figure:BPM}(a). The proposed BPM contains a domain adaptive segmentation network $G_m$ and a momentum network $M_m$ that improves the stability of pseudo labels. Let $(x_s, y_s) \in \mathcal{S}$ denote the source RGB image and its pixel-wise annotation map, $x_s\in \mathbb{R}^{H\times W\times 3}$, $y_s \in \mathbb{R}^{H\times W}$. And $(\tilde{x}_t)\in \widetilde{T}_m$ represent a purified target RGB image from the $m\tp{th}$ purified subdomain $\widetilde{T}_m$, $\tilde{x}_t \in \mathbb{R}^{H\times W\times 3}$. Note that $H$ and $W$ represent the size of height and width. Our BPM applies the mixing in two directions: $\mathcal{S}\to \widetilde{T}_m$ and $\widetilde{T}_m \to \mathcal{S}$.

On the direction of mixing from $\mathcal{S}\to \widetilde{T}_m$, we choose ClassMix~\cite{olsson2021classmix} because the source image $x_s$ has the pixel-wise annotation map $y_s$. We first randomly select some classes from $y_s$. Then, we define $\Psi\in \{0,1\}^{H \times W}$ as a binary mask in which $\Psi(h,w)=1$ when the pixel position $(h,w)$ of $x_s$ belongs to the selected classes, and $\Psi(h,w)=0$ otherwise. While ClassMix suggests directly copying the corresponding pixels of selected classes of $x_s$ onto $\tilde{x}_t$, the mixed image generated by ClassMix contains inconsistent style distribution which might hinder the adaptation performance. To cope with the limitation, the proposed BPM applies photometric transform $\Gamma$ on the selected source pixels to the style of target image before directly copying them onto it. Let $\Psi \odot x_s$ represent the selected source pixels by the mask $\Psi$, and $\odot$ is element-wise multiplication. We first calculate the histograms of selected source pixels in LAB color space, and match them with $\{\widetilde{H}_m^l, \widetilde{H}_m^a, \widetilde{H}_m^b \}$. The translated source pixels is represented as $\Gamma(\Psi \odot x_s)$. Then, we copy the translated source pixels onto $\tilde{x}_t$. We present some qualitative results in Figure~\ref{figure:mixexp}. Note that no ground-truth annotation is available for $\tilde{x}_t$. Thus, we send the purified target image $\tilde{x}_t$ to the momentum network $M_m$ to generate a stable prediction map $\tilde{y}'_t$ as the pseudo label. The mixing process on the direction of $\mathcal{S}\to \widetilde{T}_m$ by BPM is shown as
\begin{equation}
    \label{eq:mixed1}
    \begin{aligned}
        x_{\psi} &= \Gamma(\Psi \odot x_s) + (\mathbf{1} - \Psi) \odot \tilde{x}_t, \\
        y_{\psi} &= \Psi \odot y_s + (\mathbf{1} - \Psi) \odot \tilde{y}'_t,
    \end{aligned}
\end{equation}
where $x_{\psi}$ is the generated mixed image, $y_{\psi}$ is the corresponding mixed pseudo label, and $\Gamma(\cdot)$ is the photometric transform of the source selected pixels by histogram matching on LAB color space.

On the direction of mixing from $\widetilde{T}_m \to \mathcal{S}$, however, it is impossible to choose ClassMix since no ground-truth annotation is available for $\tilde{x}_t$. Inspired by CutMix~\cite{yun2019cutmix}, we generate another binary mask $\Phi \in \{0,1\}^{H\times W}$ by sampling rectangular bounding box $(d_x, d_y, d_w, d_h)$ according to the uniform distribution; $d_x\sim U(0,W), d_y \sim U(0, H), d_w=W\sqrt{1-\eta}, d_h=H\sqrt{1-\eta}$, where $\eta \sim U(0,1)$, $(H,W)$ are the height and width of the image. The binary mask $\Phi$ is formed by filling with $1$ the pixel positions inside the bounding box, and filling with $0$ other positions. With the help of $\Phi$, we select the target pixels $\Phi \odot \tilde{x}_t$ and transform them into the source style. The transformed target pixel is represented by $\Delta(\Phi \odot \tilde{x}_t)$. Then we paste them onto the source image $x_s$. We present the mixing of $\widetilde{T}_m \to \mathcal{S}$ at
\begin{equation}
    \label{eq:mixed2}
    \begin{aligned}
        x_{\phi} &=\Delta(\Phi \odot \tilde{x}_t) + (\mathbf{1} - \Phi) \odot x_s, \\
        y_{\phi} &= \Phi \odot \tilde{y}'_t + (\mathbf{1} - \Phi) \odot y_s,
    \end{aligned}
\end{equation}
where $x_{\phi}$ is the other generated mixed image, $y_{\phi}$ is the corresponding mixed pseudo label, and $\Delta(\cdot)$ is the photometric transform of the target selected pixels by histogram matching on LAB color space.

we $(x_{\psi}, y_{\psi})$ $ (x_{\phi}, y_{\phi})$ and $(x_s, y_s)$ to train the segmentation network $G_m$ and the momentum network $M_m$. We first optimize the parameters of $G_m$ through
\begin{equation}
\label{eq:bpm}
\begin{aligned}
        \mathcal{L}_{BGM}(\theta_m) = \sum_{\forall x_s \in S} \sum_{\forall \tilde{x}_t \in \widetilde{T}_m}
        \Big[ \mathcal{L}_{CE} \Big(G_m(x_s), y_s\Big)  & + \alpha \mathcal{L}_{CE} \Big(G_m(x_{\psi}), y_{\psi}\Big) \\
         &+ \beta \mathcal{L}_{CE} \Big(G_m(x_{\phi}), y_{\phi}\Big) \Big]
\end{aligned},
\end{equation}
where $\theta_m$ represent the parameters of $G_m$, $\mathcal{L}_{CE}$ is the cross-entropy loss for the predicted segmentation maps and the ground-truth or pseudo labels, $\alpha$ and $\beta$ are the hyper-parameters to control the effect of the mixing of both the directions for the loss function. To help the momentum network $M_m$ provide stable pseudo labels, we update the parameters of $M_m$, represented by ${\theta}'_m$, using an exponential moving average (EMA) with a momentum $\lambda \in [0,1]$. After finishing the training iteration $t$, ${\theta}'_m$ is updated by 
\begin{equation}
    {\theta'_m}^{t+1} = \lambda {\theta'_m}^{t} + (1-\lambda)\theta_m.
\end{equation}

\subsection{Multi-teacher Adaptive Knowledge Distillation}
We propose a multi-teacher framework followed by an adaptive knowledge distillation to align the domain gaps between the source domain and all the target subdomains. Given a purified subdomain $\widetilde{T}_m$, we adopt a BPM as a specific teacher model to minimize the domain gap between $\mathcal{S}$ and $\widetilde{T}_m$. And we train the proposed multi-teacher framework by minimizing the loss function $\mathcal{L}_{MT}$ on all the teacher models, \ie
\begin{equation}
    \mathcal{L}_{MT} = \sum_{m=1}^{k^*}\mathcal{L}_{BGM}(\theta_m),
\end{equation}
where $\mathcal{L}_{BGM}(\theta_m)$ (defined in Equation~\ref{eq:bpm}) is the loss function of the segmentation network $G_m$ in the $m\tp{th}$ teacher model, and $k^*$ is the optimal number of the subdomains. Moreover, We learn a segmentation network $G_{sd}$ as the student network via an adaptive knowledge distillation from all the teacher networks $\{G_m: 1\leq m\leq k^*\}$. Given a random target data from $x_t \in \mathcal{T}$, we send $x_t$ to all the teachers model, and the student is to learn from a weighted average of the all teacher's predictions $O_w(x_t)$, based on the teacher's confidence score. We adopt the entropy of $G_m$'s prediction map $G_m(x_t)\in \mathbb{R}^{H\times W \times C}$ as the confidence of the $m\tp{th}$ teacher model, where $C$ is the total number of classes we consider. Thus, the weight $w_m$ for the $m\tp{th}$ teacher and the average prediction $G_{out}(x_t)$ are formulated as 
\begin{equation}
    \begin{aligned}
        & w_{m} = \frac{\sum_{h,w,c} G_{m}(x_t)\log\big[G_m(x_t)\big]}{\sum_{m'}\sum_{h,w,c} G_{m'}(x_t)\log\big[G_{m'}(x_t)\big]},\\
        & G_{out}(x_t) = \sum_{m=1}^{k^*} w_m G_{m}(x_t).
    \end{aligned}
\end{equation}
On this basis, we optimize the student segmentation network $G_{sd}$ with a distillation loss $\mathcal{L}_{D}$ defined by
\begin{equation}
    \mathcal{L}_{D} =\sum_{x_t \in \mathcal{T}} \mathcal{L}_{KL}\Big[G_{sd}(x_t) || G_{out}(x_t) \Big],
\end{equation}
where $\mathcal{L}_{KL}$ is KL divergence loss function between the output of $G_{sd}$ and $G_{out}$. The goal of the multi-teacher adaptive knowledge distillation is to achieve the optimal parameters ${\theta_{sd}}^*$  of the student segmentation network $G_{sd}$ by
\begin{equation}
    {\theta_{sd}}^* = \min_{\theta_{sd}} \mathcal{L}_{MT} + \mathcal{L}_{D}.
\end{equation}
\noindent \textbf{Online Updating with Consistency Regularization.} To evaluate the generalization of our approach, we directly evaluate our student network on the open domains as shown in Table~\ref{table:opengta5} and Table~\ref{table:opensynthia}. Additionally, after finishing the compound domain adaptation training, we also provide a fast and short online updating for the student network using consistency regularization. This would further boost the generalization of the student network. Given an RGB image $x_o$ from an open domain, we first match the style of $x_o$ to other standard styles from the existing target subdomains. The standard styles are defined as the mean histograms $\{\widetilde{H}_m^l, \widetilde{H}_m^a, \widetilde{H}_m^b \}$ (defined in~\ref{SSP}). The newly transformed images are $\{x_o^m;1\leq m\leq k^* \}$, where $x_o^m$ is generated by matching $x_o$ to the style of the $m\tp{th}$ subdomain $\widetilde{T}_m$. Thus, we conduct an online updating for the student network $G_{sd}$ by
\begin{equation}
    \min_{\theta_{sd}} \sum_{m=1}^{k^*} \mathcal{L}_1 \big(G_{sd}(x_o^m), G_{sd}(x_o)\big),
\end{equation}
where $\mathcal{L}_1$ is the mean absolute loss. After the online updating, we test the student network with newly learnt parameters again on the open domains.

\section{Experiments}

\subsection{Experimental Setup}
\subsubsection{Dataset.} In this work, we adopt the synthetic datasets, including GTA5~\cite{richter2016playing} and SYNTHIA~\cite{RosCVPR16} as the source domains. GTA5 contains $24,966$ annotated images of $1,914 \times 1,052$ resolution. SYNTHIA consists of $9,400$ images with $1,280 \times 760$ resolution. Furthermore, we adopt C-Driving~\cite{liu2020open} as the compound target domains which contains real images of $1,280 \times 720$ resolution collected from different weather conditions. Following the settings of previous works~\cite{liu2020open,park2021discover,gong2021cluster}, we use the $14,697$ rainy, snowy, cloudy images as the compound target domain and adopt $627$ overcast images as the open domain. We also use ACDC~\cite{sakaridis2021acdc} as another compound target domain and the evaluation results are shown in supplementary material.
We further adopt Cityscapes~\cite{cordts2016cityscapes}, KITTI~\cite{abu2018augmented}, and WildDash~\cite{zendel2018wilddash} as the open domains to evaluate the generalization ability of the proposed approach.

\subsubsection{Implementation Details.} 
We adopt DeepLab-V2~\cite{chen2017deeplab} with ResNet101 backbone~\cite{he2016deep} pre-trained on ImageNet~\cite{deng2009imagenet}. All the images from target domain are rescaled into $1,280 \times 720$ and then randomly cropped into $640 \times 360$. The batch size is set up with $2$ and the total number of training iterations is $2.5\times 10^{5}$. We adopt stochastic gradient descent to optimize all the segmentation networks, with a weight decay of $5\times 10^{-4}$ and momentum of $0.9$. The learning rate is set up with an initial value of $2.5\times 10^{-4}$ and decreased by polynomial decay with an exponent of $0.9$. The momentum network has the same network architecture as the segmentation network. Existing mixing techniques contain CutMix~\cite{yun2019cutmix}, CowMix~\cite{french2020milking} and ClassMix~\cite{olsson2021classmix}. We adopt ClassMix on the mixing direction of the source domain to the target domain, and we apply CutMix on the mixing direction of the target domain to the source domain. Both $\alpha$ and $\beta$ are set up with $1$ in the experiments. To increase the robustness of the segmentation model, we adopt data augmentations, including flipping, color jittering, and Gaussian blurring on the mixed images.

\begin{table*}[t!]
\caption{The performance comparison of mean IoU on the compound domain. Our approach is compared with the state-of-the-art UDA and OCDA approaches on (a) GTA5$\rightarrow$C-Driving and (b) SYNTHIA$\rightarrow$C-Driving benchmark dataset with ResNet-101 as the backbone. Note that mIoU$^{11}$ represents the mean IoU of 11 classes, excluding the class with $^{*}$.}   
\label{table:c_driving}
\centering
\resizebox{\textwidth}{!}{
\begin{tabular}{l|c|c c c c c c c c c c c c c c c c c c c|c}
\multicolumn{21}{c}{ (a) GTA5$\to$C-Driving}\\
\toprule[1.0pt]
Method & Type & \rotatebox{90}{road} & \rotatebox{90}{sidewalk} & \rotatebox{90}{building} & \rotatebox{90}{wall} & \rotatebox{90}{fence} & \rotatebox{90}{pole} & \rotatebox{90}{light} & \rotatebox{90}{sign} & \rotatebox{90}{veg} & \rotatebox{90}{terrain} & \rotatebox{90}{sky} & \rotatebox{90}{person} & \rotatebox{90}{rider} & \rotatebox{90}{car}& \rotatebox{90}{truck} & \rotatebox{90}{bus} & \rotatebox{90}{train} & \rotatebox{90}{mbike} & \rotatebox{90}{bike} & mIoU \\
\hline
Source & -  & 73.4 & 12.5 & 62.8 & 6.0 & 15.8 & 19.4 & 10.9 & 21.1 & 54.6 & 13.9 & 76.7 & 34.5 & 12.4 & 68.1 & 31.0 & 12.8 & 0.0 & 10.1 & 1.9 & 28.3 \\
CDAS~\cite{liu2020open} & OCDA  & 79.1 & 9.4 & 67.2 & 12.3 & 15.0 & 20.1 & 14.8 & 23.8 & 65.0 & 22.9 & 82.6 & 40.4 & 7.2 & 73.0 & 27.1 & 18.3 & 0.0 & 16.1 & 1.5 & 31.4 \\
CSFU~\cite{gong2021cluster} & OCDA  & 80.1 & 12.2 & 70.8 & 9.4 & 24.5 & 22.8 & 19.1 & 30.3 & 68.5 & 28.9 & 82.7 & 47.0 & 16.4 & 79.9 & 36.6 & 18.8 & 0.0 & 13.5 & 1.4 & 34.9 \\
SAC~\cite{araslanov2021self} & UDA & 81.5 & 23.8 & 72.0 & 10.3 & 27.8 & 23.0 & 18.2 & 34.1 & 70.3 & 27.9 & 87.8 & 45.0 & 16.9 & 77.6 & 38.5 & 19.8 & 0.0 & 14.0 & 2.7 & 36.4 \\
DACS~\cite{tranheden2021dacs}& UDA  & 81.9 & 24.0 & 72.2 & 11.9 & 28.6 & 24.2 & 18.3 & 35.4 & \textbf{71.8} & 28.0 & 87.7 & 44.9 & 15.6 & 78.4 & 39.1 & \textbf{24.9} & 0.1 & 6.9 & 1.9 & 36.6 \\
DHA\cite{park2021discover} & OCDA & 79.9 & 14.5 & 71.4 & \textbf{13.1} & 32.0 & \textbf{27.1} & 20.7 & 35.3 & 70.5 & 27.5 & 86.4 & 47.3 & 23.3 & 77.6 & 44.0 & 18.0 & 0.1 & 13.7 & 2.5 & 37.1 \\
\hline
\rowcolor{Gray} Ours & OCDA  & \textbf{85.3} & \textbf{26.2} & \textbf{72.8} & 10.6 & \textbf{33.1} & 26.9 & \textbf{24.6} & \textbf{39.4} & 70.8 & \textbf{32.5} & \textbf{87.9} & \textbf{47.6} & \textbf{29.2} & \textbf{84.8} & \textbf{46.0} & 22.8 & \textbf{0.2} & \textbf{16.7} & \textbf{5.8} & \textbf{40.2} \\
\bottomrule
\end{tabular}}

\resizebox{\textwidth}{!}{
\begin{tabular}{l|c|c c c c c c c c c c c c c c c c|c|c}
\multicolumn{19}{c}{ (b) SYNTHIA$\to$C-Driving}\\
\toprule[1.0pt]
Method & Type & \rotatebox{90}{road} & \rotatebox{90}{sidewalk} & \rotatebox{90}{building} & \rotatebox{90}{wall} & \rotatebox{90}{fence} & \rotatebox{90}{pole} & \rotatebox{90}{light} & \rotatebox{90}{sign$^{*}$} & \rotatebox{90}{veg} & \rotatebox{90}{sky} & \rotatebox{90}{person} & \rotatebox{90}{rider$^{*}$} & \rotatebox{90}{car}&  \rotatebox{90}{bus$^{*}$} & \rotatebox{90}{mbike$^{*}$} & \rotatebox{90}{bike$^{*}$} & mIoU$^{16}$ & mIoU$^{11}$\\
\hline
Source  & - & 33.9 & 11.9 & 42.5 & 1.5 & 0.0 & 14.7 & 0.0 & 1.3 & 56.8 & 76.5 & 13.3 & 7.4 & 57.8 & 12.5 & 2.1 & 1.6 & 20.9 & 28.1  \\
CDAS~\cite{liu2020open} & OCDA  & 54.5 & 13.0 & 53.9 & 0.8 & 0.0 & 18.2 & 13.0 & 13.2 & \textbf{60.0} & 78.9 & 17.6 & 3.1 & 64.2 & 12.2 & 2.1 & 1.5 & 25.3 & 34.0 \\
CSFU~\cite{gong2021cluster} & OCDA & 69.6 & 12.2 & 50.9 & 1.3 & 0.0 & 16.7 & 12.1 & 13.6 & 56.2 & 75.8 & 20.0 & 4.8 & 68.2 & 14.1 & 0.9 & 1.2 & 26.1 & 34.8 \\
SAC~\cite{araslanov2021self} & UDA & 69.8 & 13.4 & 56.2 & 1.7 & 0.0 & 20.0 & 9.6 & 13.7 & 52.5 & 78.1 & 29.1 & 15.5 & 68.9 & 10.9 & \textbf{3.2} & 1.2 & 27.7 & 36.3 \\
DACS~\cite{tranheden2021dacs} & UDA &62.1 & \textbf{15.2} & 48.8 & 0.3 & 0.0 & 19.7 & 10.3 & 9.6 & 57.8 & \textbf{84.4} & 35.2 & 18.9 & 67.8 & 16.0 & 2.2 & 1.7 & 28.1 & 36.5 \\
DHA~\cite{park2021discover} & OCDA &67.5 & 2.5 & 54.6 & 0.2 & 0.0 & \textbf{25.8} & 13.4 & \textbf{27.1} & 58.0 & 83.9 & 36.0 & 6.1 & 71.6 & \textbf{28.9} & 2.2 & 1.8 & 29.9 & 37.6  \\
\rowcolor{Gray}Ours & OCDA  & \textbf{73.4} & \textbf{15.2} & \textbf{57.1} & \textbf{1.8} &  0.0  & 23.2 & \textbf{13.5} & 23.9 & 59.9 & 83.3 & \textbf{40.3} & \textbf{22.3} & \textbf{72.2} & 23.3 &  2.3 & \textbf{2.2} &  \textbf{32.1} & \textbf{40.0} \\
\bottomrule
\end{tabular}}
\end{table*}

\subsection{Results}
To demonstrate the efficacy of our approach, we conduct experiments on the benchmark datasets of GTA5$\to$C-Driving and SYNTHIA$\to$C-Driving. We first compare our approach with the existing state-of-the-art OCDA approaches: CDAS~\cite{liu2020open}, DHA~\cite{park2021discover}, and CSFU~\cite{gong2021cluster}. Furthermore, we compare the proposed approach with the current state-of-the-art UDA approaches SAC~\cite{araslanov2021self} and DACS~\cite{tranheden2021dacs}.

\subsubsection{Compound Domain Adaptation.} We first compare the performance of our approach with existing state-of-the-art OCDA and UDA approaches on GTA5 $\rightarrow$C-Driving, shown in Table~\ref{table:c_driving}a. All the results are generated on the validation set of C-Driving. Training only with the source data leads to $28.3\%$ of mean IoU over the 19 classes. As the first work in OCDA, CDAS achieves $31.4\%$ on the mean IoU of all the classes. CSFU generates $34.9\%$ of mean IoU, and DHA produces $37.1\%$ of mean IoU. This is because both CSFU and DHA adopt the subdomain separation step and GAN framework, and DHA uses a more effective multi-discriminator to minimize the domain gaps. In comparison, the latest UDA approaches DACS and SAC show $36.6\%$ and $36.4\%$, outperforming both CDAS and CSFU. The reason behind is that both DACS and SAC adopt various self-supervision techniques to minimize the domain gaps, which proves to be more effective than GAN-based approaches. In comparison, the proposed approach demonstrates effectiveness on this benchmark dataset with $40.2\%$ of mean IoU over all classes. 

We present experimental results on SYNTHIA$\rightarrow$C-Driving shown in Table~\ref{table:c_driving}b. We consider the 11 classes for final evaluation. The proposed method achieves $40.0\%$ of mean IoU over the 11 classes. For other OCDA approaches, DHA achieves $37.6\%$, CSFU produces $34.8\%$, and CDAS generates $34.0\%$ of mean IoU. Moreover, the UDA approaches DACS and SAC generate $36.5\%$ and $36.3\%$ of mean IoU. Our approach outperforms all the existing OCDA approaches and the latest UDA approaches.
\begin{table}[t]
    \label{table:opendms}
    \caption{The comparison of mean IoU on the open domains. The domain generalization (DG) model is trained only with the source domain. All the models are tested on the validation set of C-Driving Open (O), cityscapes (C), KITTI (K), and wildDash (W). We also present the scores of our approach without online updating (w/o Updating) and with online updating (w/ Updating).}
    \begin{subtable}[t]{0.49\textwidth}
        \centering
        \caption{GTA5 as the source domain.}
        \resizebox{\textwidth}{!}{
        \begin{tabular}{c|c|c c c c |c}
            \toprule
            \multicolumn{7}{c}{GTA5} \\
            \hline
            Method & Type & O & C & K & W & Avg \\
            \hline
            CSFU~\cite{gong2021cluster} & OCDA & 38.9 & 38.6 & 37.9 & 29.1 & 36.1 \\
            DACS~\cite{tranheden2021dacs} & UDA & 39.7 &  37.0 & 40.2 & 30.7 & 36.9 \\
            RobustNet~\cite{choi2021robustnet} & DG & 38.1 & 38.3 & 40.5 & 30.8 & 37.0 \\
            DHC~\cite{park2021discover} & OCDA & 39.4 & 38.8 & 40.1 & 30.9 & 37.5 \\
            \hline
            Ours (w/o Updating)    & OCDA & 41.8 & 40.9 & 44.0 & 32.9 & 40.0 \\
            \rowcolor{Gray} Ours (w/ Updating)     & OCDA & 42.5 & 41.7 & 44.3 & 34.6 & 40.8 \\
            \bottomrule
            \end{tabular}
            }
      \label{table:opengta5}
    \end{subtable}
    \hfill
    \begin{subtable}[t]{0.49\textwidth}
        \centering
        \caption{SYNTHIA as the source domain.}
        \resizebox{\textwidth}{!}{
        \begin{tabular}{c|c|c c c c |c}
            \toprule
            \multicolumn{7}{c}{SYNTHIA} \\
            \hline
            Method & Type & O & C & K & W & Avg \\
            \hline
            CSFU~\cite{gong2021cluster} & OCDA & 36.2 & 34.9 & 32.4 & 27.6 & 32.8 \\
            DACS~\cite{tranheden2021dacs} & UDA & 36.8 &  37.0 & 37.4 & 28.8 & 35.0 \\
            RobustNet~\cite{choi2021robustnet} & DG & 37.1 & 38.3 & 40.1 & 29.6 & 36.3\\
            DHC~\cite{park2021discover} & OCDA & 38.9 & 38.0 & 40.6 & 30.0 & 36.9 \\
            \hline
            Ours (w/o Updating)    & OCDA & 41.5 & 40.3 & 42.7 & 30.1 & 38.7 \\
            \rowcolor{Gray} Ours (w/ Updating)     & OCDA & 42.6 & 41.1 & 43.4 & 30.9 & 39.5 \\
            \bottomrule
            \end{tabular}
        }
        \label{table:opensynthia}
     \end{subtable}
\end{table}

\subsubsection{Generalization to the Open Domains.} We also evaluate the domain generalization of the proposed approach against existing UDA and OCDA approaches. The results are presented in Table~\ref{table:opengta5} and~\ref{table:opensynthia}. Our work is compared with the latest domain generalization (DG) approach RobustNet~\cite{choi2021robustnet}. For all the UDA and OCDA approaches, we first train them with the labeled source and the unlabeled target images, and we evaluate their performance with the validation of the open domains. RobustNet generates $37.0\%$ of mean IoU in Table~\ref{table:opengta5} and  $36.3\%$ of mean IoU in Table~\ref{table:opensynthia}. Note that RobustNet only requires labeled source data during training. This shows that the DG approach is more effective in generalizing to the open domains than the existing UDA and OCDA approaches DACS and CSFU. Without any online updating, our approach achieves $40.0\%$ of mean IoU in Table~\ref{table:opengta5} and $38.7\%$ of mean IoU in Table~\ref{table:opensynthia}. Our approach outperforms all the UDA approaches, OCDA approaches, and the DG approach listed in the table. The reason might be that our approach is more powerful for learning the domain invariant features which improve the generalization of the model toward novel domains. The performance gain of our approach with updating further shows the efficacy of the proposed online updating with consistency regularization.

\subsection{Ablation Study}
\subsubsection{Generating Optimal Subdomains.}
\begin{figure}[t!]
    \centering
    \includegraphics[width=\textwidth]{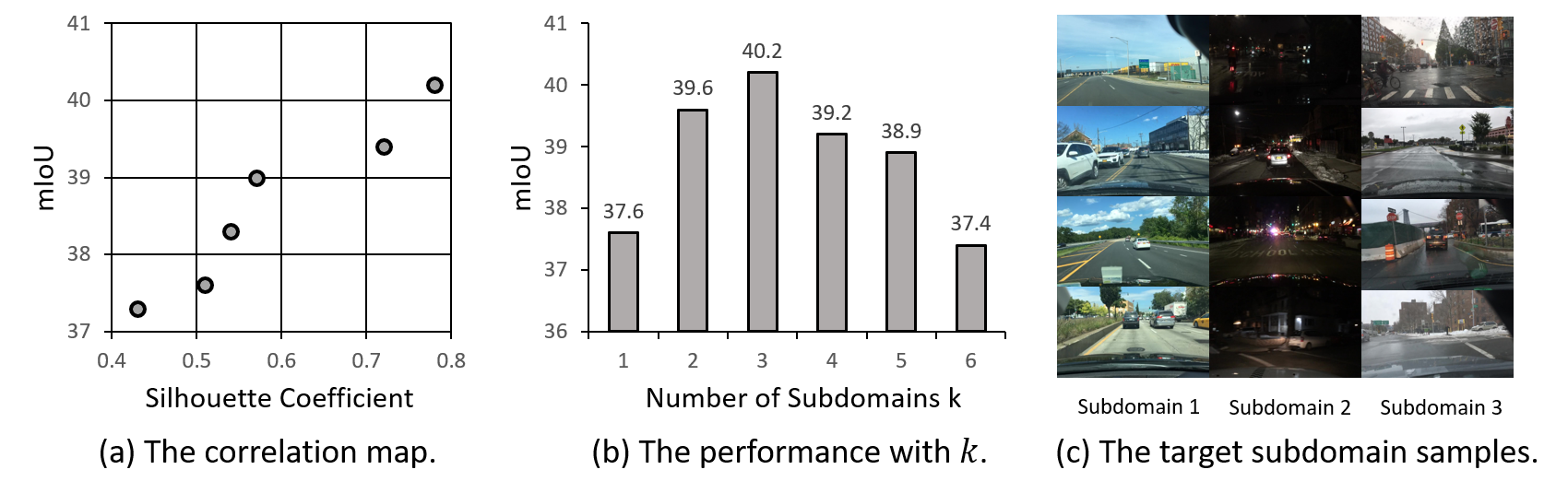}
    \caption{We conduct the ablation study on the proposed automatic domain separation using GTA5$\to$C-Driving with ResNet101 backbone. (a) The scatterplot shows the correlation between our approach's mean IoU and the Silhouette Coefficient score. (b) The mean IoU of our approach with different number of subdomains $k$. (c) The sample images from the subdomains of the C-Driving dataset.}
    \label{figure:ass}
\end{figure}
We first conduct the ablation study on the correlation between the mean IoU of the proposed approach with Silhouette Coefficient (SC) score on the subdomain separation in Figure~\ref{figure:ass}(a). It shows a positive correlation, which means that the SC score is effectively finds the optimal number of subdomains for the compound target domain. Moreover, we evaluate the mean IoU score with the different number of subdomains $k$ in Figure~\ref{figure:ass}(b). Finally, we set up $k=3$ and present the sample images from the subdomains of the C-Driving dataset in Figure~\ref{figure:ass}(c). We also evaluate the efficacy of the proposed subdomain style purification (SSP) in Table~\ref{table:drop}. Without using SSP, the performance drops $0.5\%$ of mean IoU.
\begin{table}[t!]
    \caption{The ablation study on the efficacy of the components of our model. (a) We compare with one baseline model DACS~\cite{tranheden2021dacs} and evaluate the performance gain of the bidirectional photometric mixing and the multi-teacher learning. (b) We evaluate the performance drop of our model by removing each component from it. Our model is trained GTA5$\to$C-Driving with ResNet101 backbone and tested on C-Driving validation set.}
    \begin{subtable}[t]{0.47\textwidth}
        \centering
        \caption{The performance gain.}
        \resizebox{\textwidth}{!}{
        \begin{tabular}{l|c}
                    \toprule
                    \multicolumn{2}{c}{GTA5$\rightarrow$C-Driving}  \\
                    \hline
                    Model & mIoU \\
                    \hline
                    DACS~\cite{tranheden2021dacs} & 36.6 \\
                    DACS + Multi-teacher Learning & 39.1 \\
                    DACS + Bidirectional Mixing  & 37.3 \\
                    DACS + Photometric Mixing ($\Gamma,\Delta$) & 37.4 \\
                    DACS + Bidirectional Photometric Mixing & 37.8 \\
                    \hline
                     Ours & 40.2 \\
                    \bottomrule
        \end{tabular}
            }
      \label{table:gain}
    \end{subtable}
    \hfill
    \begin{subtable}[t]{0.47\textwidth}
        \centering
        \caption{The performance drop.}
        \resizebox{\textwidth}{!}{
        \begin{tabular}{l|c c}
                    \toprule
                    \multicolumn{3}{c}{GTA5$\rightarrow$C-Driving}  \\
                    \hline
                    Configuration  & mIoU & Gap \\
                    \hline
                    w/o Multi-teacher Learning & 38.0 & -2.2\\
                    w/o Mixing on One Direction ($\alpha=0$)& 38.5 & -1.7 \\
                    w/o Mixing on One Direction ($\beta=0$) & 38.9 & -1.3 \\
                    w/o Subdomain Style Purification & 39.7 &  -0.5 \\
                    w/o Adaptive Distillation  & 39.6 & -0.6  \\
                    \hline
                     Full Framework & 40.2 & - \\
                    \bottomrule
        \end{tabular}
        }
        \label{table:drop}
     \end{subtable}
\end{table}
\subsubsection{Multi-teacher and Single Model.} 
\begin{figure}[t!]
    \centering
    \includegraphics[width=0.95\textwidth]{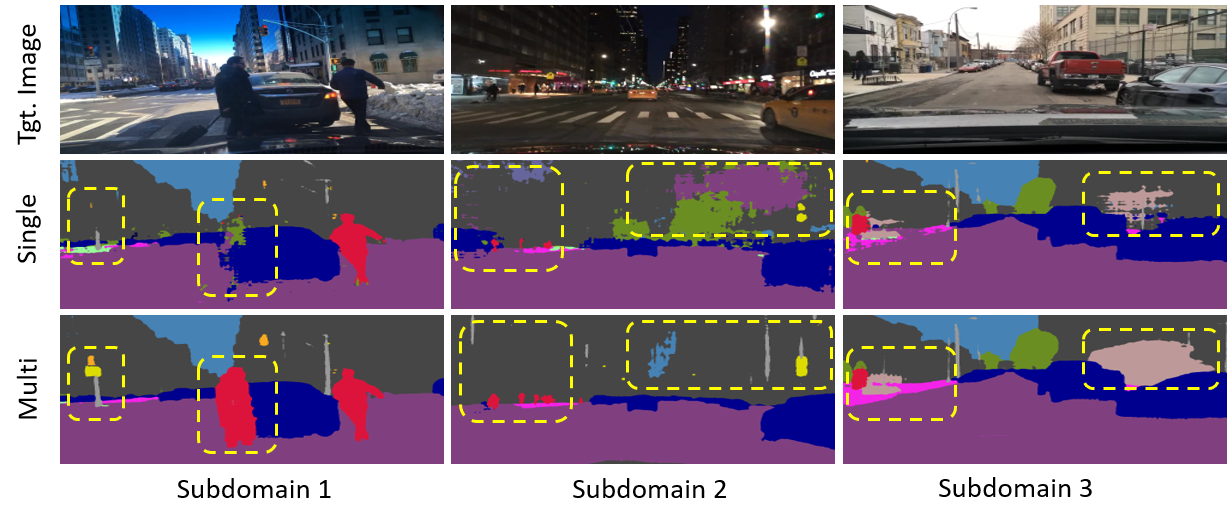}
    \caption{We present the predicted segmentation maps of the target images from every target subdomain. The maps in the second row are generated using a single model. The maps in the third row are generated using the multi-teacher models.}
    \label{figure:subsub.png}
\end{figure}

The ablation study on the multi-teacher learning of our proposed approach is presented in Table~\ref{table:gain} and Table~\ref{table:drop}. Applying a single model in our approach delivers $38.0\%$ of mean IoU, leading to the the most significant drop $2.2\%$, shown in Table~\ref{table:drop}. We further combine DACS with multi-teacher learning, and the mean IoU reaches from $36.6\%$ to $39.1\%$. We argue that utilizing a single model is less effective than the multi-teacher models. Because adapting to one subdomain might hinder the single model from adapting to other dissimilar subdomains. Thus, we employ a multi-teacher framework in which each teacher adapts to one subdomain separately. And the multiple teachers together provide a comprehensive guide to the student model to adapt to all the target subdomains. We further present the qualitative results about the target image prediction maps from each subdomain by the multi-teachers and the single-teacher model in Figure~\ref{figure:subsub.png}. 
\begin{figure}[t!]
    \centering
    \includegraphics[width=0.95\textwidth]{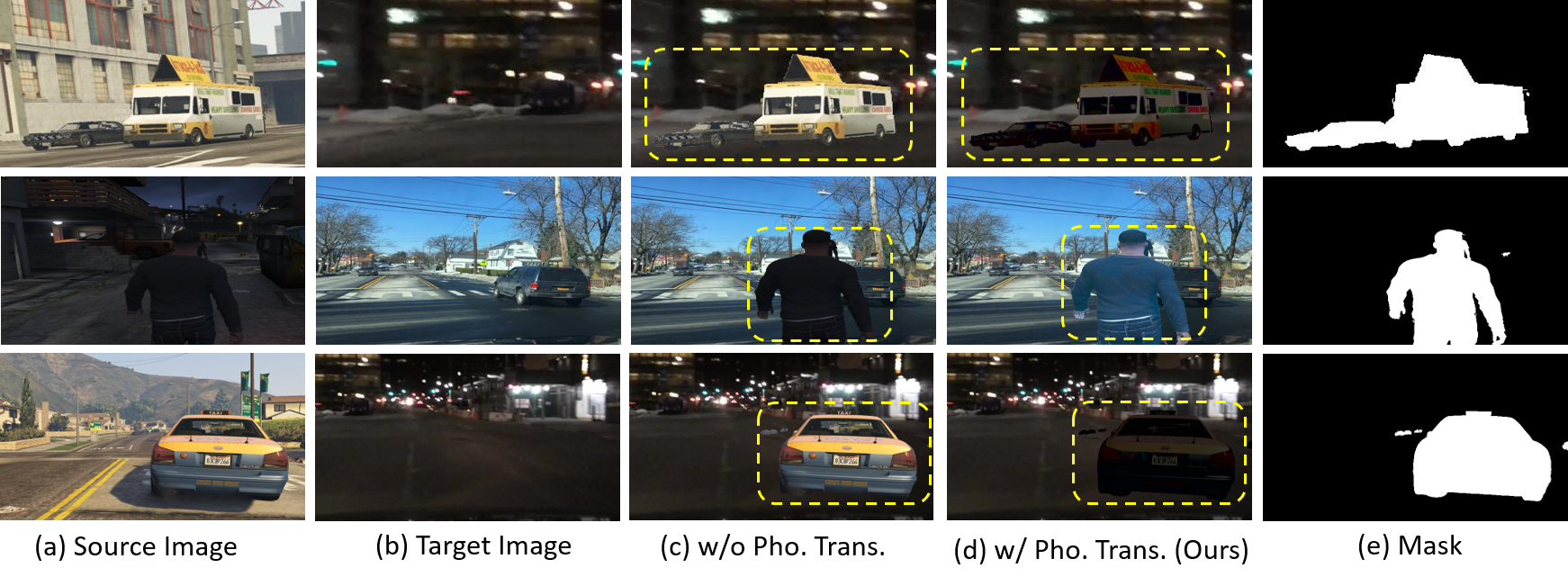}
    \caption{We compare the mixed images from the source domain to the target domain. (a) the source image; (b) the target image; (c) the mixed images without using photometric transform, and the style inconsistency exists; (d) the mixed images using photometric transform, and the style inconsistency is mitigated; (e) the mask to crop the source image.}
    \label{figure:mixexp}
\end{figure}

\subsubsection{Bidirectional Photometric Mixing.}
We further conduct the ablation study for the bidirectional photometric mixing (BPM), shown in Table~\ref{table:gain} and Table~\ref{table:drop}. Our model is trained on GTA5 $\to$ C-Driving with ResNet101 backbone and tested on C-Driving validation set. By making $\alpha=0$ to remove the mixing on one direction (ClassMix), the mean IoU drops $1.7\%$, while making $\beta=0$ to remove the other directional mixing (CutMix), it decreases by $1.3\%$. This suggests that ClassMix contributes slightly more to the final performance. We also use the baseline model DACS for an in-depth analysis. We add the bidirectional photometric mixing with the DACS, the performance increase from $36.6\%$ to $37.8\%$ shown in Table~\ref{table:gain}; we then combine DACS with only bidirectional mixing, the mean IoU rise up to $37.3\%$; we further add DACS with only photometric transform on mixing (use $\Gamma$ and $\Delta$), the mean IoU reaches to $37.4\%$. The reason behind is that DACS utilizes a simple mixing method that contains only one direction and generates the mixed image with the style inconsistency inside. However, we propose a bidirectional mixing 
scheme and apply the photometric transform to mitigate the style inconsistency on the generated images. We present the qualitative results to show this issue in Figure~\ref{figure:mixexp}. The style inconsistency is mitigated in Figure~\ref{figure:mixexp}(d) compared with Figure~\ref{figure:mixexp}(c) on the mixing direction from the source domain to the target domain.
\section{Conclusion}
Open compound domain adaptation (OCDA) considers the target domain as the compound of multiple unknown subdomains. In this work, we first propose automatic domain separation to find the optimal number of subdomains. Then we design a multi-teacher framework with bidirectional photometric mixing to align the domain gap between the source domain and the compound target domain, and we further evaluate its generalization to novel domains. Our current work is only focused on segmentation task and we leave the study on other visual tasks for future research.

\section{Acknowledgment}
This work was supported by the Korea Institute of Energy Technology Evaluation and Planning (KETEP) and the Ministry of Trade, Industry \& Energy (MOTIE) of the Republic of Korea (No. 20224000000100).

\clearpage
%
%
\bibliographystyle{splncs04}
\bibliography{egbib}

\clearpage

\pagestyle{headings}
\mainmatter
\def\ECCVSubNumber{6380}  

\title{Supplemental Material:\\
ML-BPM: Multi-teacher Learning with Bidirectional Photometric Mixing for Open Compound Domain Adaptation in Semantic Segmentation} 

\titlerunning{ECCV-22 submission ID \ECCVSubNumber} 
\authorrunning{ECCV-22 submission ID \ECCVSubNumber} 
\author{Anonymous ECCV submission}
\institute{Paper ID \ECCVSubNumber}

\titlerunning{ML-BPM}
%
\author{Fei Pan\inst{1} \and
Sungsu Hur\inst{1} \and
Seokju Lee\inst{2} \and
Junsik Kim\inst{3} \and
In So Kweon\inst{1}
}
\authorrunning{F. Pan et al.}
%
\institute{KAIST, South Korea. \email{\{feipan, sshuh1215, iskweon77\}@kaist.ac.kr}\and
KENTECH, South Korea. \email{slee@kentech.ac.kr} \and
Harvard University, USA. \email{mibastro@gmail.com}}
\maketitle

\section{Subdomain Style Purification and the t-SNE Visualization}
\begin{figure}
    \centering
    \includegraphics[width=\textwidth]{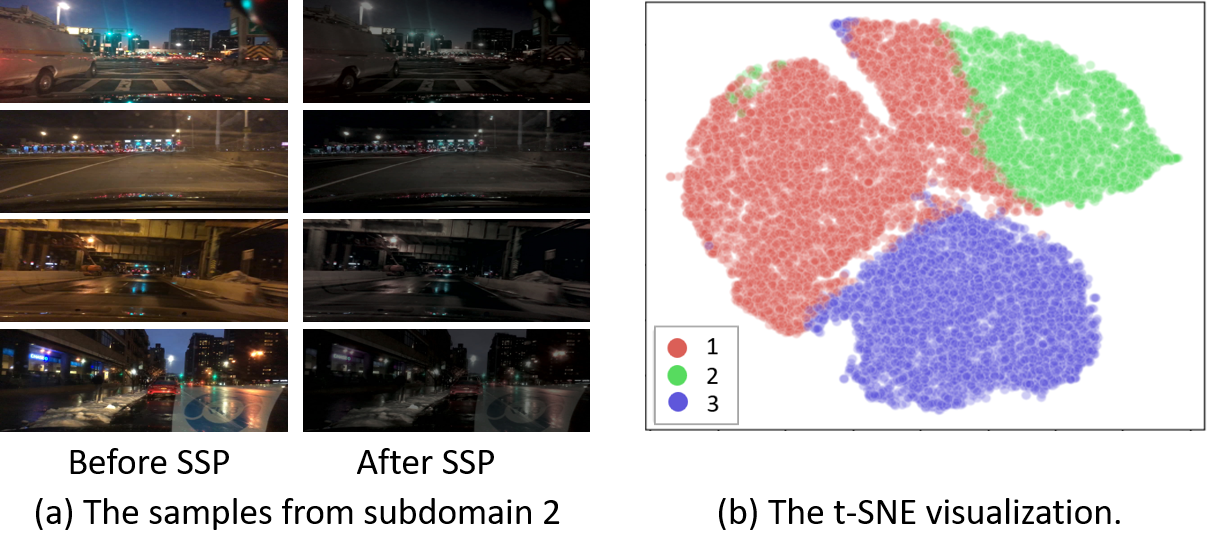}
    \caption{(a) presents the noisy samples from Subdomain 2 of C-Driving dataset before subdomain style purification (before SSP) and after subdomain style purification (after SSP). (b) shows the t-SNE visualization of the concatenated histograms of the C-Driving dataset on LAB color space when $k=3$.} 
    \label{figure:ssp}
\end{figure}

As mentioned in Section 3.2, it is hard to guarantee that the images from the same target subdomain have the same style. In other words, small domain gaps might still results from the various image styles in each subdomain. We propose subdomain style purification to unify the styles of the target data that belongs to the same subdomain so that the domain gaps in these images could be further reduced. We provide the visualization of the sample images transformed by subdomain style purification (SSP) from subdomain 2 in Figure~\ref{figure:ssp} (a). Note that the images from \emph{before SSP} in Figure~\ref{figure:ssp}(a) has the styles different from the \emph{standard style}, and they are transformed into the standard style with the help of histogram matching on the LAB color space. We further set up $k=3$ and present the t-SNE visualization of the concatenated histograms of the C-Driving images from LAB color space.

\noindent \textbf{The reason of subdomain style purification (SSP).} With the help of automatic domain separation, the number of abnormal samples with different styles is small. Though these abnormal samples might be helpful for the model's generalization, they could also lead to a negative transfer, which further hinders the model from learning domain invariant features in a specific subdomain. With GTA5$\to$C-Driving, we get a $0.5\%$ of mIoU drop on average over all the subdomains without using SSP, as shown in Table 3(b).

\section{ACDC Dataset}
We also evaluate the proposed approach on another ACDC dataset[24]. ACDC dataset contains real-world images from the road scenes in diverse weather conditions, including fog, nighttime, rain and snow. We consider the $2,800$ images of fog, nighttime and rain from the training split of ACDC as the compound domain; the $400$ snow images with pixel-wise annotations of ACDC training split are taken as the open domain. The final performance is evaluated on the validation set of ACDC, which contains $306$ images with ground-truth maps.  

\begin{table*}[h]
\caption{The performance comparison of mean IoU on the compound target domain (fog, nighttime, and rain) and the open domain (fog) of ACDC. Our approach is compared with the state-of-the-art UDA and OCDA approaches on (a) GTA5$\rightarrow$ACDC and (b) SYNTHIA$\rightarrow$ACDC benchmark dataset with ResNet-101 as the backbone.}   
\label{table:acdc}
\centering
\resizebox{\textwidth}{!}{
\begin{tabular}{l|c|c c c c c c c c c c c c c c c c c c c|c|c}
\multicolumn{23}{c}{ (a) GTA5$\to$ACDC}\\
\toprule[1.0pt]
\multicolumn{1}{c|}{} & \multicolumn{1}{c|}{}  & \multicolumn{20}{c|}{Compound} & Open \\ \cline{3-23}
\multicolumn{1}{c|}{Method} & \multicolumn{1}{c|}{Type} & \rotatebox{90}{road} & \rotatebox{90}{sidewalk} & \rotatebox{90}{building} & \rotatebox{90}{wall} & \rotatebox{90}{fence} & \rotatebox{90}{pole} & \rotatebox{90}{light} & \rotatebox{90}{sign} & \rotatebox{90}{veg} & \rotatebox{90}{terrain} & \rotatebox{90}{sky} & \rotatebox{90}{person} & \rotatebox{90}{rider} & \rotatebox{90}{car}& \rotatebox{90}{truck} & \rotatebox{90}{bus} & \rotatebox{90}{train} & \rotatebox{90}{mbike} & \rotatebox{90}{bike} & mIoU & mIoU\\
\hline
Source & -  & 43.6 & 2.5 & 46.2 & 5.2 & 0.1 & 30.3 & 15.3 & 16.3 & 56.9 & 0.0 & 71.5 & 16.3 & 13.7 & 51.4 & 0.0 & 15.1 & 0.0 & 1.4 & 4.2 & 20.5  & 27.1\\
CDAS[13] & OCDA  & \textbf{53.2} & 5.9 & 56.1 & 10.1 & 2.6 & 22.0 & 37.1 & 11.4 & 53.9 & 23.5 & 71.3 & 27.6 & 14.6 & 47.5 & 16.8 & 19.5 & 0.0 & 3.2 & 3.8 & 25.3 & 29.1 \\
CSFU[8] & OCDA  & 47.0 & 4.1 & 53.0 & 13.9 & 1.0 & 23.2 & 41.2 & 18.8 & 55.8 & 23.2 & 72.1 & 31.5 & 10.8 & 69.1 & \textbf{26.4} & 27.8 & 0.2 & 1.7 & 2.6 & 27.6 & 30.5\\
SAC[2] & UDA & 42.6 & 4.2 & 57.6 & 11.9 & 3.8 & 23.0 & 49.7 & 23.8 & 63.6 & 31.9 & \textbf{76.0} & 30.3 & 10.5 & 65.3 & 23.6 & 23.1 & 0.1 & 0.7 & 3.2 & 28.7 & 33.6 \\
DACS[24] & UDA  & 48.9 & \textbf{9.7} & 54.5 & 16.8 & 5.7 & 22.7 & 42.0 & 22.9 & 61.3 & 29.7 & 73.7 & \textbf{32.2} & 11.6 & 63.3 & 23.2 & 26.5 & 0.0 & 1.2 & \textbf{5.2} & 29.0 & 34.8 \\
DHA[19] & OCDA & 49.8 & 5.2 & \textbf{59.1} & 10.2 & 3.1 & 25.6 & 47.8 & \textbf{27.9} & 65.1 & 32.0 & 75.2 & 29.0 & 12.2 & 61.5 & 20.5 & \textbf{32.4} & 0.0 & 1.0 & 2.0 & 29.5 & 37.5\\
\hline
\rowcolor{Gray} Ours & OCDA  & 48.4 & 5.0 & 58.2 & \textbf{25.3} & \textbf{10.0} & \textbf{35.1} & \textbf{50.4} & 26.7 & \textbf{66.8} & \textbf{33.3} & 75.8 & 32.1 & \textbf{16.7} & \textbf{73.5} & 16.8 & 26.6 & \textbf{0.2} & \textbf{3.9} & 4.6 & \textbf{32.1} & \textbf{41.6} \\
\bottomrule
\end{tabular}}

\resizebox{\textwidth}{!}{
\begin{tabular}{l|c|c c c c c c c c c c c c c c c c|c|c}
\multicolumn{20}{c}{ (b) SYNTHIA$\to$ACDC}\\
\toprule[1.0pt]
\multicolumn{1}{c|}{} & \multicolumn{1}{c|}{}  & \multicolumn{17}{c|}{Compound} & Open \\ \cline{3-20}
Method & Type & \rotatebox{90}{road} & \rotatebox{90}{sidewalk} & \rotatebox{90}{building} & \rotatebox{90}{wall} & \rotatebox{90}{fence} & \rotatebox{90}{pole} & \rotatebox{90}{light} & \rotatebox{90}{sign} & \rotatebox{90}{veg} & \rotatebox{90}{sky} & \rotatebox{90}{person} & \rotatebox{90}{rider} & \rotatebox{90}{car}&  \rotatebox{90}{bus} & \rotatebox{90}{mbike} & \rotatebox{90}{bike} & mIoU$^{16}$  &  mIoU$^{16}$ \\
\hline
Source  & - & 45.2 & 0.2 & 36.7 & 1.7 & 0.6 & 25.7 & 4.0 & 5.6 & 46.6 & 64.3 & 16.9 & 11.3 & 39.6 & 16.5 & 0.6 & 1.9 & 19.8 & 20.5  \\
CDAS[13] & OCDA  & 61.3 & 0.7 & 60.1 & \textbf{11.7} & \textbf{1.8} & 28.4 & 18.8 & 23.5 & 48.6 & 28.9 & 16.5 & 15.9 & 69.2 & 18.4 & 5.4 & \textbf{5.6} & 25.9 & 23.3  \\
CSFU[8] & OCDA & 62.6 & 0.3 & 60.3 & 8.6 & \textbf{1.8} & 21.3 & 20.7 & 29.1 & 44.5 & 22.1 & \textbf{34.5} & 19.0 & 71.1 & \textbf{23.2} & 4.4 & 4.3  & 26.7 & 24.8 \\
SAC[2] & UDA & \textbf{69.8} & 0.4 & 56.2 & 1.7 & 0.0 & 20.0 & 12.6 & 13.7 & 52.5 & \textbf{78.1} & 29.1 & 15.5 & 68.9 & 20.9 & 3.2 & 1.2 & 27.7 & 25.4 \\
DACS[24] & UDA & 55.6 & 1.1 & 55.7 & 0.1 & 0.7 & 25.8 & \textbf{31.7} & 18.3 & 65.5 & 53.7 & 31.1 & 16.6 & 69.2 & 22.5 & 2.9 & 3.1 & 28.3 & 27.0 \\
DHA[19] & OCDA &55.5 & 1.1 & 57.2 & 0.7 & 0.8 & 26.6 & 22.7 & 24.6 & \textbf{65.8} & 58.4 & 29.6 & \textbf{23.9} & 70.8 & 19.5 & 5.4 & 4.2 & 29.2 & 27.3  \\
\rowcolor{Gray}Ours & OCDA  & 66.7 & \textbf{1.7} & \textbf{62.4} & 10.8 & 1.4 & \textbf{30.8} & 23.9 & \textbf{29.2} & 62.6 & 69.0 & 31.6 & 14.6 & \textbf{71.8} & 22.9 & \textbf{6.8} & 4.5 &  \textbf{31.9} & \textbf{29.1} \\
\bottomrule
\end{tabular}}
\end{table*} 

We present the performance comparison of mean IoU in Table~\ref{table:acdc}. For the compound target domain of ACDC (fog, nighttime, rain), we achieve $32.1\%$ of mean IoU on GTA5$\to$ACDC and $31.9\%$ of mean IoU on SYNTHAI$\to$ACDC, outperforming all the UDA and OCDA approaches in the list. We also evaluate the generalization of our approach compared with other works. After finishing the compound domain adaptation training, all the models are directly tested on the open domain of ACDC (snow). Note that the snow images have never been used in training before. Under the benchmark datasets GTA5$\to$ACDC and SYNTHIA$\to$ACDC, our approach shows $41.6\%$ and $29.1\%$ of mean IoU. This demonstrates that our approach has better generalization ability toward novel domains (snow).


%
%

\begin{table}[t!]
    \centering
    \caption{The evaluation on GTA5$\to$C-Driving.}
    \label{tab:lambda}
    \resizebox{0.5\linewidth}{!}{
\begin{tabular}{l c c c c c c}

\multicolumn{7}{c}{ (a) ImageNet pre-trained VGG-16 Backbone} \\
\toprule[1.0pt]
\multicolumn{1}{c}{\multirow{2}{*}{Method}} & \multicolumn{3}{c}{Compound (C)} & \multicolumn{1}{c}{Open (O)} & \multicolumn{2}{c}{Average} \\ \cline{2-7} 
\multicolumn{1}{c}{}  & Rainy & Snowy & Cloudy & Overcast & C & C+O  \\ \hline
CDAS[13]      &    23.8       &   25.3      &     29.1      &   31.0    &       26.1      &    27.3          \\
CSFU[8]       &    24.5     &   27.5     &    30.1    &   31.4    &     27.7        &    29.4       \\
DACS [24]     &    26.8     &   29.2     &    35.1    &   35.9    &     30.4        &    31.8       \\
DHA[19]       &    27.1     &   30.4     &    35.5    &   36.1    &     32.0        &    32.3      \\
\hline
\rowcolor{Gray} Ours      &    34.5     &   35.8       &    39.9       &    40.1   & \textbf{36.7}  & \textbf{37.5}   \\
\bottomrule
\end{tabular}}
\resizebox{0.5\linewidth}{!}{
\begin{tabular}{c|c c c c}
     \multicolumn{5}{c}{ (b) Mixing Algorithm Comparison} \\
     \toprule[1.0pt]
     Algorithm & BPM (\emph{Ours}) & ClassMix [15] & CutMix [31] & CowMix [7] \\
     \hline
     mIoU & \textbf{40.2} & 39.1 & 37.6 & 37.4 \\
     \bottomrule
\end{tabular}
}
\end{table}

\section{The Practicability of Our Approach}
Though we use the multi-teacher models for training, our approach still has strong practicability for the two following reasons: these teacher models are trained simultaneously; only a single student model from distillation is needed for inference. The size of the student model is not affected by the number of the subdomains. With the number of the subdomains $k^*$, the FLOPS and the number of parameters of our \emph{multi-teacher's} model are $327.08\times 10^9$ and $43.8\times k^* \times 10^6$. After the adaptive knowledge distillation, the FLOPS and number of parameters of our \emph{student} model is $327.08\times 10^9$ and $43.8\times 10^6$. \\

\noindent \textbf{The VGG-16 backbone and different mixup algorithms.} We use VGG-16 backbone network for evaluation. The experimental results on GTA5$\to$C-Driving in Table~\ref{tab:lambda}(a) demonstrates the effectiveness of our approach against existing works with ImageNet pre-trained VGG-16 as the backbone. We provide the comparison to existing domain mixup algorithms in the same setting, including ClassMix[15], CutMix[31], and CowMix[7]. \\

\noindent \textbf{The online updating on the open domains.} Our online updating is conducted on each sample from the open domain, thus it is still domain generalization at the testing stage. Our student model $G_{sd}$ is trained through the adaptive distillation from all the subdomain's segmentation models $\{G_m\}_{m=1}^{k^*}$ (Eq. (10, 11)). Each $G_m$ is optimized by Eq. (7) with the help of the mean teacher $M_m$, following the work of DACS[24]. We also used $M_m$ instead of $G_m$ for distillation but do not see significant performance gain. \\

\noindent \textbf{The reason of using bidirectional mixing. The performance of using pseudo-labels of target data for ClassMix.} Using the photometric transform $\Delta$ (Eq.(6)) on target-to-source (t2s) mixing, we enforce the consistency of prediction between the target and the mixed image, which are taken as additional augmentation to improve the model's performance (Table.3(a,b)). With the experiment on GTA5$\to$C-Driving, we get $40.1\%$ of mIoU on using pseudo-labels of target data for ClassMix on target-to-source mixing, similar to ours $40.2\%$ (Table 1(a)). Table~\ref{tab:lambda} (b) shows that our BPM outperforms existing mixing algorithms ClassMix [15], CutMix [31], and CowMix [7].

\end{document}